\documentclass[letterpaper, 10 pt, conference]{ieeeconf}  % Comment this line out if you need a4paper

\IEEEoverridecommandlockouts % This command is only needed if you want to use the \thanks command
\usepackage{amsmath,amssymb,amsfonts}
\usepackage{graphicx}
\usepackage{xcolor}
\usepackage[hyphens]{url}
\usepackage[hidelinks,pdftex]{hyperref}
\usepackage[T1]{fontenc}
\usepackage[all]{hypcap}

\usepackage{siunitx}
\usepackage{float}
\usepackage{tikz}
\usepackage{pgfplots}
\pgfplotsset{compat=1.18}

\definecolor{SB_RED}{RGB}{189,64,36}
\definecolor{SB_BLUE}{RGB}{47,119,182}
\definecolor{SB_LIGHTRED}{RGB}{239,208,201}
\definecolor{SB_LIGHTBLUE}{RGB}{204,221,237}

\graphicspath{{./graphics/}}

\begin{document}
% \title{\bf Learning to Grasp on the Moon from 3D Octree Observations\\with Deep Reinforcement Learning}
% \title{\bf Two-stage Learning Process for Reinforcement Learning-Based Mapless Collision-free Navigation on a Mars Rover}

% \title{\bf Two-stage Learning Process for Mapless Collision-free\\Navigation on a Mars Rover with Deep Reinforcement Learning}

% \title{\bf Teacher-Student Policy Learning for Mapless Collision-free\\Navigation on a Mars Rover with Deep Reinforcement Learning}

% \title{\bf Teacher-Student Policy Learning for Mapless Navigation on a\\Planetary Rover with End-To-End Deep Reinforcement Learning}

\title{\bf Teacher-Student Reinforcement Learning for Mapless Navigation using a Planetary Space Rover}

%% Authors (names)
\author{Anton~Bjørndahl~Mortensen\textsuperscript{1}, Emil~Tribler~Pedersen\textsuperscript{1}, Laia~Vives~Benedicto\textsuperscript{1}, Lionel~Burg\textsuperscript{1}, \\Mads~Rossen~Madsen\textsuperscript{1}, Simon~Bøgh\textsuperscript{1,2}}

\maketitle
\thispagestyle{empty}
\pagestyle{empty}

%% Authors (affiliations and emails)
\footnotetext[1]{Robotics \& Automation Group, Department of Materials and Production, Aalborg University, Denmark. {\tt \{abmo19}, {\tt etpe19}, {\tt lvives22}, {\tt lburg22}, {\tt mrma19\}@student.aau.dk, \tt sb@mp.aau.dk}}
\footnotetext[2]{AAU SPACE, \url{https://www.space.aau.dk}, Aalborg University, Denmark.}

\hypersetup{%
	pdfinfo ={%
			Title={Two-stage Learning Process for RL-Based Mapless Collision Free Navigation on a Mars Rover},%
			Subject={2022 IEEE/RSJ International Conference on Intelligent Robots and Systems (IROS)},%
			Author={Anton Bjørndahl Mortensen, Emil Tribler Pedersen, Laia Vives Benedicto, Lionel Burg, Mads Rossen Madsen and Simon Bøgh},%
			Creator={LaTeX},%
			Keywords={Space Robotics and Automation; Reinforcement Learning; Deep Learning in Collision-free Navigation},%
		}%
}%

%%%%%%%%%%%%%%%%%%%%%%%%%%%%%%%%%%%%%%%%%%%%%%%%%%%%%%%%%

\begin{abstract} 
% Introduction/Problem Statement:
We address the challenge of enhancing navigation autonomy for planetary space rovers using reinforcement learning (RL). The ambition of future space missions necessitates advanced autonomous navigation capabilities for rovers to meet mission objectives.
% Significance/Complexity of the Problem:
RL's potential in robotic autonomy is evident, but its reliance on simulations poses a challenge. Transferring policies to real-world scenarios often encounters the "reality gap", disrupting the transition from virtual to physical environments.
% Specific Challenges/Considerations:
The reality gap is exacerbated in the context of mapless navigation on Mars and Moon-like terrains, where unpredictable terrains and environmental factors play a significant role. Effective navigation requires a method attuned to these complexities and real-world data noise.
% Proposed Solution/Methodology:
We introduce a novel two-stage RL approach using offline noisy data. Our approach employs a teacher-student policy learning paradigm, inspired by the "learning by cheating" method. The teacher policy is trained in simulation. Subsequently, the student policy is trained on noisy data, aiming to mimic the teacher's behaviors while being more robust to real-world uncertainties.
% Implementation/Results:
Our policies are transferred to a custom-designed rover for real-world testing. Comparative analyses between the teacher and student policies reveal that our approach offers improved behavioral performance, heightened noise resilience, and more effective sim-to-real transfer.

Videos, simulation environment, source code and datasets:
\href{https://github.com/abmoRobotics/isaac_rover_2.0}{\url{https://github.com/abmoRobotics/isaac_rover_2.0}}, 
\href{https://github.com/abmoRobotics/isaac_rover_2.0_learning_by_cheating}{\url{https://github.com/abmoRobotics/isaac_rover_2.0_learning_by_cheating}}

% [\textit{Source code and datasets for simulation and training will be made publicly available on github upon publication}].

\end{abstract}

%%%%%%%%%%%%%%%%%%%%%%%%%%%%%%%%%%%%%%%%%%%%%%%%%%%%%%%%%

\section{Introduction}

Enhancing the autonomy of planetary rovers is vital for efficient space exploration. NASA's Perseverance rover utilizes the Approximate Clearance Evaluation (ACE) for navigation. Although ACE is innovative, it can be overly conservative in challenging terrains~\cite{otsu2020fast}. To mitigate this, the advanced Probabilistic ACE (p-ACE) was developed~\cite{ghosh2018probabilistic}. This work pivots towards exploring Reinforcement Learning (RL) for mapless navigation, capitalizing on RL's adaptability and capability to derive optimal strategies, offering a promising approach for navigating unknown environments~\cite{9611016}. Recent GPU-based simulator advancements suggest faster training periods, minimizing the need for laborious hyperparameter adjustments~\cite{makoviychuk2021isaac}. Yet, these simulators, while efficient, approximate real-world dynamics, leading to a "reality gap". Transferring policies from simulation to the actual world may face performance issues~\cite{zhao2020sim}.

\begin{figure}[t]
	% \vspace{2.057mm}
	\centering
	\includegraphics[width=1.0\linewidth]{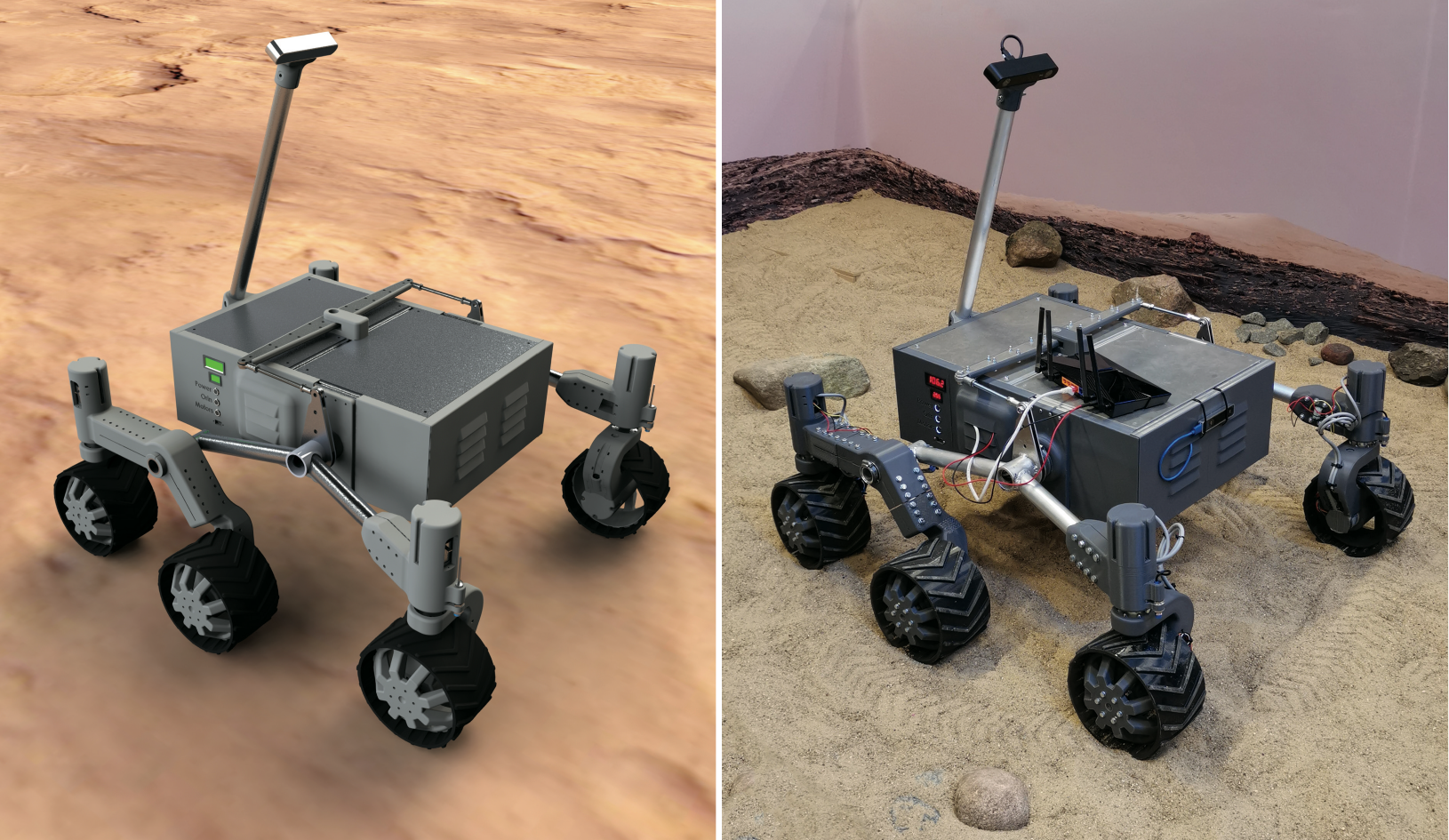}
    \vspace{-5mm}
	\caption{We employ a model of the rover in simulation and its physical counterpart to investigate the agent's ability to navigate a Mars-like environment without prior mapping and to assess its sim-to-real transference.}
	\label{fig:rover_sim_and_real}
	\vspace{-1.0\baselineskip}
\end{figure}

Reinforcement learning with 3D observations on space rovers has enabled robust strategies for tasks like moon rock grasping \cite{orsula2022learning}. Accurate rover localization is crucial for planetary navigation. Even high-resolution satellite imagery, like from the HiRISE camera, has limitations in resolution and accuracy \cite{wu2019absolute}. A central challenge for space rovers is autonomous navigation in unfamiliar terrains. Mapless navigation, relying solely on local 3D observations, offers a solution, contrasting with methods like SLAM that require local map construction. By prioritizing real-time decisions over extensive mapping, mapless navigation reduces computational demands, emphasizing the need for its development to boost rover autonomy in upcoming missions.

In this paper, we develop a two-stage end-to-end deep reinforcement learning approach for mapless navigation on planetary rovers across challenging Martian-like terrains. As depicted in Figure~\ref{fig:rover_sim_and_real}, our simulation framework offers a comprehensive exploration of the agent's ability to navigate these terrains without the need for prior mapping. This not only enhances our understanding of the agent's capabilities but also provides a foundation for assessing its sim-to-real transfer potential. To ensure the practical applicability of our approach, we test the sim-to-real performance on a robot that mirrors the kinematics of NASA's Perseverance rover.

\subsection{Related Work}\label{sec:related_work}

Enhancing the autonomy of planetary rovers has been a focal point in space exploration. The unique challenges posed by extraterrestrial terrains, especially in the absence of comprehensive data sources like GPS and high-resolution satellite imagery, necessitate innovative solutions. This section delves into the pertinent literature, highlighting navigation techniques, the potential of reinforcement learning, and the nuances of sim-to-real transfer.

\subsubsection{Navigation in Planetary Environments}
Navigating planetary terrains, unlike Earth, presents many challenges due to the absence of pivotal data sources such as GPS, comprehensive 2D and 3D scans, and the limited knowledge about environments like Mars or the Moon~\cite{Howard2006}. The recent advancements in imaging, such as the high-resolution imagery from the HiRISE camera, have provided a resolution of 0.25 m/pixel, offering a more detailed view of the Martian surface~\cite{mcewen2007mars}. However, the inherent challenges persist. In light of this, our work introduces a mapless navigation strategy, which, rather than relying on extensive planetary data, harnesses information from a procedurally generated simulation environment.

\subsubsection{Reinforcement Learning and Reality Gap}
Reinforcement learning has been effectively applied to control autonomous vehicles in structured environments like highways, primarily for tasks such as lane keeping and lane changing~\cite{liu2019reinforcementdriving,hoel2019combining}. The potential of RL in navigating less structured environments has been investigated using various machine learning techniques~\cite{9611016,cardona2019autonomous,tai2017virtual,ruan2019mobile}. However, many of these studies have limited their evaluations to simulations without addressing the challenges of sim-to-real transfer~\cite{9611016,cardona2019autonomous,ruan2019mobile}. Salvato et al.~\cite{9606868} highlight the difficulties in transferring policies from simulation to reality, pointing out factors like sensor noise and differences in dynamics. To mitigate these challenges, approaches such as domain randomization, adversarial RL, and transfer learning have been proposed.

\subsubsection{Sim-to-real Techniques}
Domain randomization seeks to bolster an agent's adaptability by introducing noise to its input, thereby narrowing the reality gap~\cite{vuong2019pick}. Salvato et al.\cite{9606868} introduce adversarial RL and transfer learning as alternative strategies. While adversarial RL pushes the agent to adapt to increasingly challenging scenarios, transfer learning emphasizes continued training on the physical system. However, the latter remains impractical for planetary rovers due to the risks associated with extraterrestrial deployments. Our approach, inspired by Miki et al.\cite{miki2022learning}, employs a two-stage learning process, leveraging the concept of learning by cheating~\cite{chen2020learningbycheating}. This methodology, although prevalent in control frameworks~\cite{miki2022learning,lee2020learning,de2022depth}, remains relatively unexplored in navigation. Our work pioneers its application in planetary mapless navigation, emphasizing challenges unique to terrains like Mars or the Moon.

\subsection{Contributions}
The central contributions of this work is the design and development of a two-stage reinforcement learning framework tailored for mapless navigation on planetary rovers. This two-stage approach accelerates the training process, seamlessly integrating the precision of policies trained with optimal data with the resilience of those trained using domain randomization. Our primary objective is to craft robust navigation policies capable of handling diverse terrains with obstacles such as holes, valleys, and rocks of varying dimensions. Recognizing the inherent challenges of real-world operations, our methodology leans heavily on a meticulously crafted, parallelized, physics-based 3D simulation, enabling efficient data collection from multiple robots simultaneously.
We introduce a GPU-accelerated simulation environment in NVIDIA Isaac Sim tailored for planetary settings, pivotal in learning mapless navigation policies across varied terrains. The real-world applicability of these policies is directly affected by the simulation's realistic physics and its capability to emulate diverse terrain conditions. 
Furthermore, our work employ a novel blend of online and offline data for swifter, improved agent training. Employing a teacher-student model, the teacher, with access to privileged data, aids in training the student. This method outperforms merely adding more domain randomization during teacher training.
We validate the proposed methodology in a successful demonstration of mapless navigation training and its zero-shot sim-to-real transfer and testing on our real-world rover.
\begin{figure*}[t!]
    \centering % 
    \includegraphics[width=\textwidth]{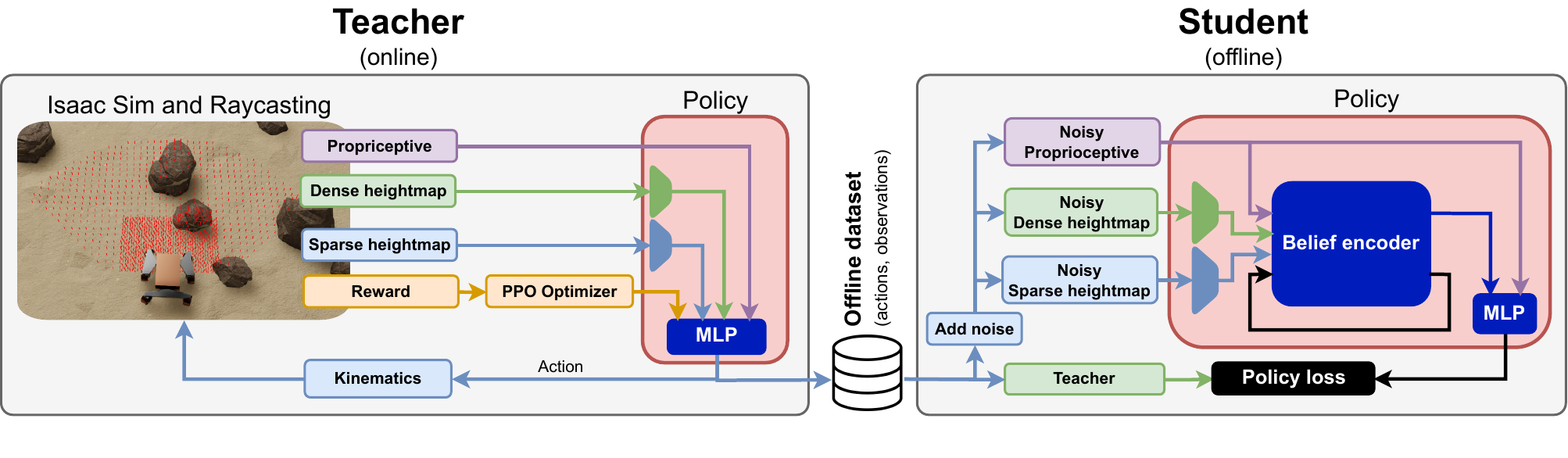}
    \vspace{-10mm}
    \caption{The two-stage framework entails a student policy that learns to imitate the teacher, given noisy and non-privileged information. On the left is the Reinforcement learning loop for training the teacher agent, where the policy gets observations extracted from the simulated environment. The rover then performs an action based on the output of the policy and the kinematics. On the right is the student model, which takes the saved actions and observations from the teacher policy and feeds them to the belief encoder.}
    \label{fig:Learning_by_Cheating_overview}
    \vspace{-3mm}
\end{figure*}

\section{Extraterrestrial Mapless Navigation} 
\label{sec:methods}

Our approach incorporates a teacher-student framework, offline data generation, and end-to-end learning of extraterrestrial mapless navigation in a procedurally-generated physics simulation environment. An overview of the approach is presented in \autoref{fig:Learning_by_Cheating_overview}.

\subsection{Teacher-Student Framework}
Our two-stage teacher-student framework is used for training a teacher and student agent shown in \autoref{fig:Learning_by_Cheating_overview}. In the first stage, a teacher policy is trained with privileged information in the form of observations without noise. The student policy is then trained to imitate the teacher, and is given noisy input to make the policy more robust.

\subsection{Action and Observation Space} \label{sec:observations}

The action space of the agent is defined as $\langle v_{lin}, v_{ang}  \rangle$, where $v_{lin}$ denotes the linear velocity, and $v_{ang}$ is the angular velocity. The linear and angular velocities are used to control the rover using Ackermann steering. Additionally, if the ratio between the velocities is below 0.15 the rover will turn on the spot. This turn-on-the-spot mode allows the agent to be more flexible when navigating in very constrained environments. 

The observation space of the agent is defined as
\begin{equation}
    o_{t}^{teacher}=\bigl\{o_t^p, o_t^e\bigl\},
\end{equation} 
where $o_t^p$ is the proprioceptive and $o_t^e$ is the exteroceptive input. The proprioceptive input is defined as

\begin{equation}\label{eq:observation}
    o_t^p = \bigl\{d(x,y), \theta_{goal},  a_{i,t-1}\bigl\},
\end{equation}

where $d(x,y)$ is the euclidean distance from the rover to the target, $\theta_{goal}$ is the angle between the rover and the target,
and $a_{i,t}$ is the $i^{th}$ action at time $t$. Furthermore, the exteroceptive input is defined as
\begin{equation}\label{eq:exteroceptive}
    o_t^e = \bigl\{ o_t^{d}, o_t^{s}\bigl\},
\end{equation}

where $o_t^{d}$ is a dense heightmap near the rover, and $o_t^{s}$ is a sparse heightmap including points up to \SI{4}{\meter} from the rover. The sample distance between each point is \SI{5}{\centi\meter} for the dense heightmap $o_t^{d}$ and \SI{15}{\centi\meter} for the sparse heightmap $o_t^{s}$. 

\subsection{Composite Reward Function}\label{subsec:rewards}
A reward function that encourages reaching the target goal without colliding with obstacles is defined as

\begin{equation}\label{eq:total_reward}
    r_{total} = r_d + r_a + r_v + r_h ,
\end{equation}

where the individual terms are presented below. The main goal of the policy is to decrease the distance to the target until the target is reached. This behavior is encouraged by rewarding the agent for decreasing the distance to the target. This reward is formulated as 

\begin{equation}\label{eq:distance_reward}
    r_d = \frac{\omega_d}{1+\frac{1}{3}d(x,y)},
\end{equation}

where $d(x,y)$ is the euclidean distance from the rover to the target, and $\omega$ is a weight (see \autoref{tab:hyperparameters}). However, this has proved to be very hard to optimize by itself as the policy needs to output relatively steady actions to move anywhere. Consequently, a reward that penalizes oscillations of the actions is defined as

\begin{equation}\label{eq:Oscillation_constraint}
    r_a = -\omega_a \sum^2_i (\alpha_{i,t}-\alpha_{i,t-1})^2,
\end{equation}

where $\alpha_{i,t}$ is the $i^{th}$ action at time $t$. Furthermore, it was observed that the rover would learn to drive backwards early in the learning process and was thus not able to see obstacles before colliding with them. To overcome this, two penalties are introduced, the first penalty punishes driving backwards which is formulated as

\begin{equation}\label{eq:Velocity_constraint}
    r_v = 
    \begin{cases}
    v_{lin} < 0    & -\omega_v \cdot \vert v_{lin} \vert\\
    v_{lin} \geq 0 & 0,
    \end{cases}
\end{equation}
where $v_{lin}$ is the linear velocity. The second penalty penalizes the agent if the angle between the target and the direction of the rover is larger than $\SI{115}{\degree}$, and is defined as 

\begin{equation}\label{eq:heading_constraint}
    r_h = 
    \begin{cases}
    \vert \theta_{goal}\vert > \SI{115}{\degree}, & -\omega_h \cdot \vert \theta_{goal} \vert \\
    \vert \theta_{goal}\vert \leq \SI{115}{\degree}, & 0.
    \end{cases}
\end{equation}

\subsection{Teacher Network Architecture}
As training is performed using an on-policy method, the policy $\pi_\theta$ is modeled using a Gaussian model.
The network architecture is designed to generate two distinct latent representations; firstly, a dense representation of the terrain in close proximity to the rover denoted $l_t^d$, and secondly, a sparse representation of the terrain denoted $l_t^s$. These are defined as
\begin{align} 
l_t^d &=  e_d\bigl(o_t^d\bigl), \\ 
l_t^s &=  e_s\bigl(o_t^s\bigl),
\end{align}
where $e_d$ and $e_s$ are encoders for the dense and sparse input. Both encoders consist of two linear layers of size [60, 20] and utilize LeakyReLU\cite{maas2013rectifier} as activation function. A multilayer perceptron is then applied to the latent representations and the proprioceptive input as
\begin{equation}\label{eq:proproceptive}
    a = mlp\bigl(o_t^p, l_t^d, l_t^s\bigl),
\end{equation}
where $a$ refers to the actions and $mlp$ is a multilayer perceptron with three linear layers of size [512, 256, 128] and LeakyReLU as activation function.

\begin{figure}[H]
    \centering
    \includegraphics[width=0.95\linewidth]{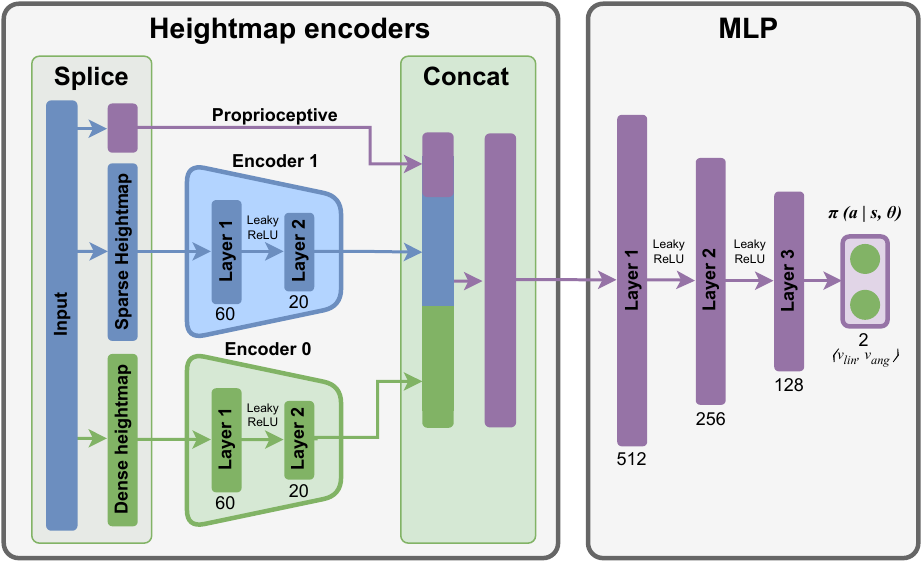}
    \caption{Network architecture for the teacher policy where the input of the dense and sparse heightmap is encoded, then concatenated and output for the MLP.}
    \label{fig:architecture_overview}
\end{figure}

\subsection{Data Collection} \label{sec:data_collection}
Data for training the student policy is obtained through the execution of the trained teacher policy in inference mode, wherein observations (distance to goal, heading to goal, previous actions, and heightmap), in addition to the inferred actions, are logged. The resulting dataset is devoid of any noise. To train the student policy we logged the observations and actions of 512 rovers over 1500 timesteps. 
This approach to data acquisition enables the student policy to be trained in a supervised fashion, wherein the teacher policy serves as the source of knowledge for the student policy. The teacher policy, having already undergone extensive training, possesses a high degree of proficiency in the task at hand. Consequently, the student policy can be expected to learn from the teacher policy's successes and failures, resulting in a policy that is better equipped to handle the given task.

\subsection{Noise}
To reduce the effects of noise on the policy and thus narrow the sim-to-real gap, a noise model is added to the observations before training the student policy. The noise model is inspired by Miki et al.~\cite{miki2022learning}, where one of three different noise modes is applied to the exteroceptive input with a probability of $\rho$. 
The three modes add the following noise levels: small deviations ($\rho=0.6$), small deviations with an offset ($\rho=0.3$), and large deviations ($\rho=0.1$) respectively. Finally, a percentage of heightmap points is set to zero to emulate missing information in the heightmap, which could be caused by occlusions or reflections.

\subsection{Student Network Architecture}
\begin{figure}[ht]
    \vspace{2.057mm}
    \centering
    \includegraphics[width=1.0\linewidth]{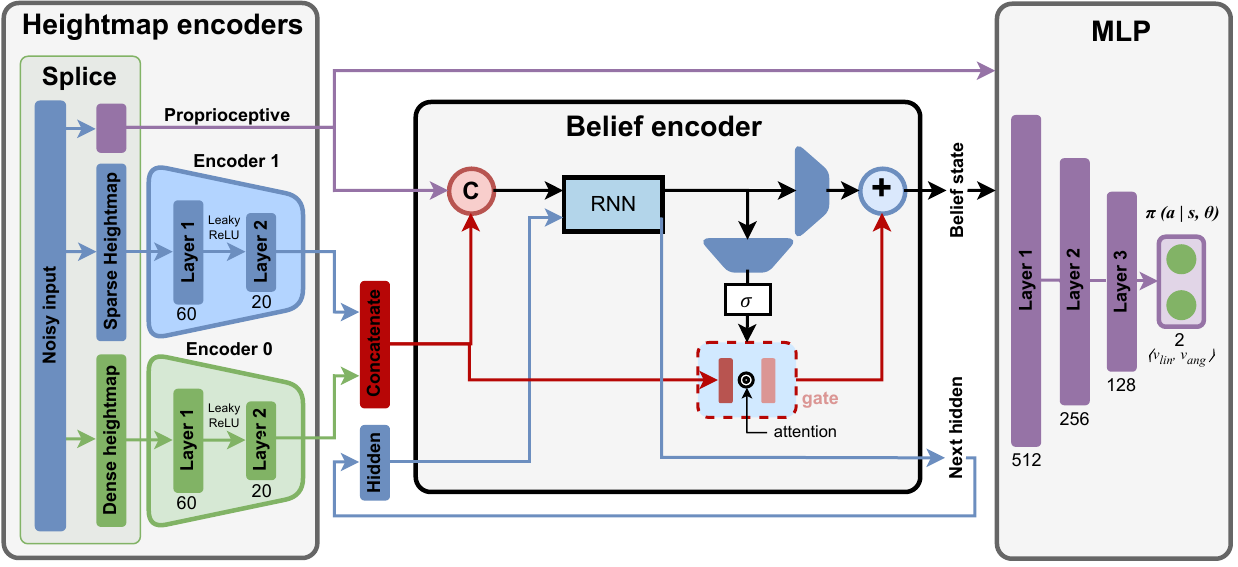}
    \vspace{-5mm}
    \caption{Network architecture for the student policy. Noise is added to the input before accessing the belief encoder, where the previous state is considered to reconstruct the ground truth and output to the MLP.}
    \label{fig:learningbycheating}
    % \vspace{-1.0\baselineskip}
\end{figure}

The student policy is modeled as a deterministic policy and uses an encoder as shown in \autoref{fig:learningbycheating}. 
Furthermore, the student policy is modeled as a time sequence, where a Recurrent Neural Network (RNN)\cite{sherstinsky2020fundamentals} is used to reconstruct the ground truth latent representations $l_t^d$ and $l_t^s$ from noisy data. This is done to imitate and remove noise present in physical systems, by using previous hidden states from the RNN. The noisy data used by the student is defined as
\begin{equation}\label{eq:student_train}
    o_t^{student} =\bigl\{o_t^p, n_e(o_t^e)\bigl\},
\end{equation}
where $n_e$ is a noise model applied to the original input.

% \begin{align*} 
% x_t', h_t &=  GRU(o_t^p, l_t^r, l_t^f, h_{t-1}) \\ 
% x_t &=  g_b(x_t') + l_t^e \cdot \sigma(g_a(b_t')) 
% \end{align*}

The belief encoder shown in \autoref{fig:learningbycheating} uses a multilayer gated recurrent unit (GRU)\cite{cho2014properties} to model the system as a time sequence, expressed as
\begin{equation}\label{eq:gated_recurrent_unit}
    x_t', h_t =  GRU(o_t^p, l_t^s, l_t^d, h_{t-1}),
\end{equation}
where $x_t'$ is an intermediate state and $h_t$ is the hidden state. Furthermore, an attention gate is applied to allow part of the latent representation to pass directly through the network. The attention gate is defined as 
\begin{equation}\label{eq:attention_unit}
    AG =  \sigma(e_a(x_t')) ,
\end{equation}
where $\sigma$ is the sigmoid activation function, and $e_a$ is an encoder with four linear layers with a size of [128, 256, 512, 1024]. Based on this the belief state is calculated as
\begin{equation}\label{eq:belief_state}
    x_t =  e_b(x_t') + [l_t^s, l_t^d] \odot AG,
\end{equation}
where $\odot$ denotes the Hadamard product and $e_b$ is another encoder with four linear layers of size [128, 256, 512, 1024]. 

Lastly, to generate actions, a multilayer perceptron with three linear layers of size [512, 256, 128] is applied to the belief state and the proprioceptive input as
\begin{equation}
    \alpha = mlp(o_t^p, x_t).
\end{equation}
The policy is optimized using the mean squared error (squared L2 norm) for the policy loss.

\section{Experiments}\label{sec:experiement_setup}
All presented methods were used to conduct a variety of experiments in simulation and on the physical rover while evaluating the sim-to-real applicability of our approach.

\subsection{Experimental Setup}
Based on the NASA Perseverance rover, a physical rover was designed to test the sim-to-real gap (see \autoref{fig:rover_sim_and_real}). The rover has a footprint of $1.03 \times \SI{1.05}{\meter}$, weighs \SI{40}{\kilo\gram}, and is equipped with an NVIDIA Orin for data processing. DB59 motors from Nanotec with a 62:1 gearing are used for the four steering and six driving motors, producing up to  \SI{12}{Nm} on the output shaft of each motor.
A ZED 2i camera is mounted on the rover for extracting a heightmap of the surrounding terrain corresponding to the one in simulation, hence allowing a transfer of the in-simulation trained policy to the physical rover. 
The heightmap is extracted by converting the point cloud captured by the camera into a grid and taking the maximum height of each grid cell. Positional tracking is performed using an Intel RealSense T265 tracking camera to determine relative distance and heading to the target goal.

\begin{figure}[b!]
    \vspace{-5mm}
    \centering
    \includegraphics[width=1.0\linewidth]{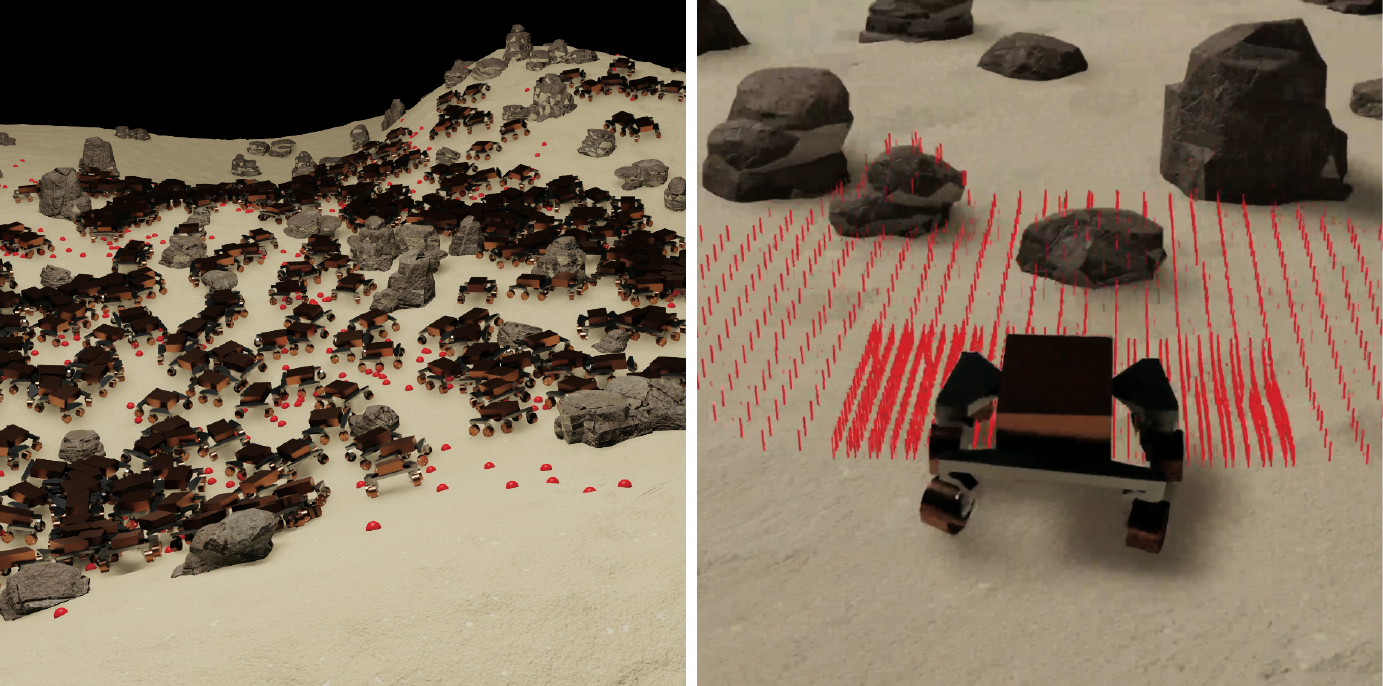}
    \vspace*{-6mm}
    \caption{Isaac Sim with 512 parallel rovers and depth map observations.}
    \label{fig:isaac_simulation}
\end{figure}

% \subsection{Simulation}
To simulate the rover and its environment, Nvidia Isaac Sim 2022.1.1 is used as seen in \autoref{fig:isaac_simulation}. The terrain used inside of the simulation is generated following a NASA terrain generation algorithm used for evaluation of the Enhanced
Autonomous Navigation of the Mars 2020 Rover~\cite{9172345}. Small bumps are generated as well as bigger hills with smoother slopes and a set of rock models is added to the resulting elevation map. Rocks are stored in different layers to facilitate ray-tracing-based collision detection.
% 
% \subsubsection{Spawns}
Rovers are spawned in a fixed grid but with randomized orientations. Goal points are spawned randomly on the perimeter of a circle of \SI{9}{\meter} radius around a rover. 
% 
%\subsubsection{Camera} Isaac Sim lacks a camera API to access data from simulated camera sensors. Therefore, the camera is emulated by a ray tracing based algorithm which calculates a height map based on the distance to each ray's first intersection point. 
% 
%\subsubsection{Collisions}\label{sec:collision} To make sure that a collision %reset is only triggered if the rover drives into a rock that it cannot climb or %drive over, there is a separate rock layer containing all impassable rocks. Ray %sources placed around the wheels and the body of the rover emit rays to detect %intersections with this layer. If an intersection occurred closer than \SI{0.25}%{\meter} to the ray source, a collision is detected. In this case, the rover is %reset to its initial spawn position, a new target is spawned, and a penalty is %applied.
% 
% \subsubsection{Collisions}\label{sec:collision} 
Collisions are detected by placing rays around the rover and checking their nearest intersection points with rocks. A separate layer for climbable rocks makes sure that the rover is only reset if it collides with a non-climbable rock.

%\subsubsection{Isaac Gym}
%The Isaac Gym API is used to conduct RL. To speed up the training process, a specifiable amount of environments is created and run in parallel, each with its own rover and goal point. The rovers of different environments do not interact with each other but they share the same terrain.

\subsection{Teacher and Student Training Procedure}\label{sec:teacher_training_process}
Training of the teacher policy takes place inside Isaac Sim, where the reinforcement learning loop visualized in \autoref{fig:Learning_by_Cheating_overview} is implemented. The training is performed using the Isaac Gym API by running 512 rovers in parallel and optimizing using Proximal Policy Optimization~(PPO)~\cite{PPO}. The implementation of the agent is done through the \textit{skrl} RL-framework \cite{serrano2022skrl}. skrl is an open-source modular library for Reinforcement Learning written in Python (using PyTorch) and supports gym-interfaces for NVIDIA Omniverse Isaac Sim/Gym \cite{makoviychuk2021isaac}, Isaac Orbit \cite{mittal2023orbit} and more. The simulation runs at  a frequency of \SI{60}{\Hz} and a controller frequency of \SI{5}{\Hz}. 

% \begin{figure}[ht]
%     \vspace{2.057mm}
%     \centering
%     \includegraphics[width=0.45\textwidth]{network500.pdf}
%     \caption{Reinforcement learning loop for training the agent, where the policy gets information from the simulation environment. The rover then performs an action based on the output of the policy and the kinematics.}
%     \label{fig:teachertraining}
%     % \vspace{-1.0\baselineskip}
% \end{figure}

The student is trained in a supervised approach using the dataset generated while training the teacher policy as described in \autoref{sec:data_collection}. It is expected to output the same action as the teacher while using noisy data as input. The training process continues for a fixed number of epochs, or until the student agent converges.
To introduce a level of robustness to the trained policy, noise is added to the collected heightmap data during the training process. This noise could take the form of sensor noise, actuator noise, or other forms of perturbations, and serves to expose the student agent to a wider range of scenarios that may arise during deployment. This, in turn, helps the student agent to generalize better and improve its overall performance.
Moreover, the student agent's performance is evaluated by measuring its success rate in reaching the target goal, as well as its efficiency in terms of the time taken to reach the goal.

\subsection{Hyperparameters}
% This section presents the hyperparameters used for training the teacher and student policies.
% \subsubsection{Teacher Parameters}
Selecting appropriate hyperparameters is a crucial step in optimizing the performance of a reinforcement learning algorithm. In the case of optimizing the teacher policy training a combination of manual and automated hyperparameter tuning was applied. An important hyperparameter is the KL threshold, which determines the maximum divergence between the new and old policy distributions. A higher KL threshold can result in faster convergence, but can also lead to instability. The teacher policy is optimized using PPO~\cite{PPO}, and the hyperparameters used for the optimizer are listed in \autoref{tab:hyperparameters}.

\begin{table}[ht]
    \centering
    \caption{Hyperparameters and reward weights used for training teacher policy. 
    % The collision penalty ($\omega_{c}$) is $10\%$ of the max score.
    }
    \label{tab:hyperparameters}
	\vspace{-0.25\baselineskip}
    \begin{tabular}{lc} 
        \hline
        \textbf{Hyperparameter} & \textbf{PPO}  \\ \hline
        Learning Rate$(l_r)$  &$ 0.0001$ \\ 
        KL Threshold$(KL)$ & $0.008$\\ 
        Horizon length & $60$ \\ 
        Discount factor ($\gamma$) & $0.99$ \\
        Lambda ($\lambda$) & $0.95$ \\ \hline
        \\ \hline
        \textbf{Reward function} & \textbf{Weight}  \\ \hline
        Relative reward ($\omega_{d}$) & $1.0$\\ 
        Oscillation Constraint ($\omega_{a}$) & $-0.01$ \\ 
        Velocity constraint ($\omega_{v}$) & $-0.005$ \\
        Heading constraint ($\omega_{h}$) & $-0.05$ \\
        Collision penalty ($\omega_{c}$) & $-0.1 \cdot \underset{\pi}{max}\text{ }\mathbb{E(\mathrm{R})} $\\ \hline
        
    \end{tabular}
	\vspace{-0.8\baselineskip}
\end{table}

% \subsubsection{Student parameters}

For the training of the student, the hyperparameters to emulate noise present in the physical world, are represented by a noise model applied to the teacher data. The noise is generated from a Gaussian distribution parameterized by $\sigma$. The individual levels of noise are shown in \autoref{tab:noise_modes}.

\begin{table}[ht]
    \centering
    \caption{Characteristics of the three noise modes added to the heightmap gathered by the teacher agent.}
    \label{tab:noise_modes}
	\vspace{-0.25\baselineskip}
    \begin{tabular}{lccc} \hline
        \textbf{Noise level} & \textbf{STD ($\sigma$)} & \textbf{offset ($\sigma$)} & \textbf{zeroed (\%)} \\ \hline
        Low          & 0.1 & 0.0  & 10\% \\
        Low + Offset & 0.1 & 0.05 & 10\% \\
        High         & 0.2 & 0.0  & 10\% \\ \hline
    \end{tabular}
	\vspace{-0.8\baselineskip}
\end{table}

% \subsection{Physical Rover}
% Based on the Perseverance rover by NASA a physical rover was constructed to test the sim-to-real gap. The rover has a footprint of $1.03 \times \SI{1.05}{\meter}$, weighs \SI{40}{\kilo\gram}, and is equipped with an NVIDIA Orin for data processing. DB59 motors from Nanotec with a 62:1 gearing are used for the steering and driving motors, producing up to  \SI{12}{\newton\meter} on the output shaft of each motor.
% Furthermore, a ZED 2i camera is mounted on the rover for extracting a heightmap of the surrounding terrain corresponding to the one in simulation, hence allowing a transfer of the in-simulation trained model to the physical rover. The heightmap is extracted by converting the point cloud captured by the camera into a grid and taking the maximum height of each grid cell. Positional tracking is performed using an Intel RealSense T265 tracking camera to determine distance and heading to the target. The design of the physical rover is shown in \autoref{fig:physical_rover}.

% \begin{figure}[H]
%     \centering
%     \includegraphics[width=0.38\textwidth]{images/Rover.png}
%     \caption{The physical rover constructed for quantifying the reality gap.}
%     \label{fig:physical_rover}
% \end{figure}

%\begin{figure}[H]
%    \vspace{2.057mm}
%    \centering
%    \includegraphics[scale=0.4]{real rover edited.png}
%    \caption{The physical rover constructed for quantifying the %reality gap.}
%    \label{fig:physical_rover}
%    \vspace{-1.0\baselineskip}
%\end{figure}

\section{Results}\label{sec:results}
To evaluate the performance of the teacher and student agents, experiments are conducted in simulation and on the physical rover.

\subsection{Simulation}
\begin{figure}[ht]
    % \vspace{2.057mm}
    \centering
    \includegraphics[width=1.0\linewidth]{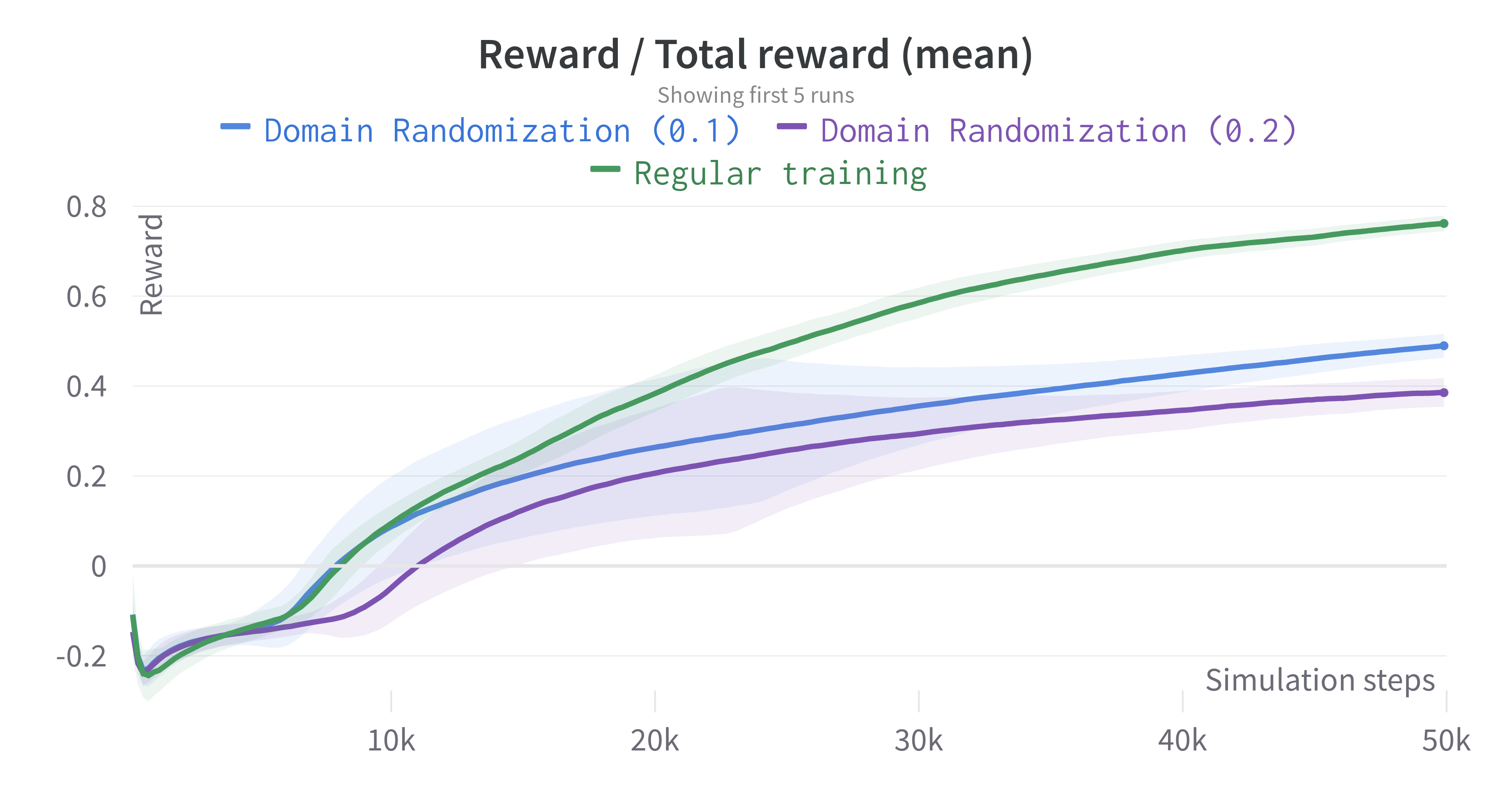}
    \vspace*{-5mm}
    \caption{Comparing regular training process without domain randomization with training with domain randomization. The graph shows the median $\pm 1\sigma$ for five runs each.}
    \label{fig:Test:Training}
\end{figure}

The first test compares the convergence rate of training a policy with and without domain randomization as shown in \autoref{fig:Test:Training}. 
The results highlight that the policy trained without domain randomization has a better convergence rate, and the policy trained with domain randomization converge to suboptimal policy.
The second test involve simulating the teacher and the student agent in inference mode in a specified environment. Their performance is evaluated using the following metrics:

\begin{itemize}
    \item Percentage of rovers reaching the goal
    \item Average duration of successfully completed episodes
    % \item Percentage of rovers getting stuck (neither reaching the goal nor colliding)
\end{itemize}

The tests are performed in a terrain with only big rocks (T1), and a more challenging terrain (T2) also including smaller rocks that can be climbed or driven over by the rover. Furthermore, Gaussian noise with a standard deviation of 0.1 is added to the heightmap and $10\%$ of the points are zeroed in some of the tests to compare the robustness of the teacher and the student policy. Results of 512 simulation episodes are shown in \autoref{tab:results_success_rate}.

% \begin{table}[ht]
%     \centering
%     \caption{Success rate and avg. episode length of agents in simulation.}
%     \label{tab:results_success_rate}
% 	\vspace{-0.25\baselineskip}
%     \begin{tabular}{lcccc} \hline
%         \textbf{Agent} & \textbf{T1} & \textbf{T1(n)} & \textbf{T2} & \textbf{T2(n)} \\ \hline
%         Teacher         & \textbf{92.6\%} & 77\% & \textbf{87\%} & 54\% \\
%          \textit{ - avg. ep. len.}    & \textit{(36.9s)} & \textit{(67.0s)} & \textit{(34.0s)} & \textit{(69.5s)}\\
%         Teacher\textsubscript{noise} & 69\% & 66\%  & 46\% & 43\%  \\
%          \textit{ - avg. ep. len.}    & \textit{(42.5s)} & \textit{(31.7s)} & \textit{(31.2s)} & \textit{(26.5s)}\\
%         Student         & 91.4\%  & \textbf{90\%}  & 85\% & \textbf{63\%}  \\ 
%          \textit{ - avg. ep. len.}    & \textit{(40.9s)} & \textit{(42.8s)} & \textit{(45.9s)} & \textit{(42.7s)}\\ \hline
%     \end{tabular}
% 	\vspace{-0.8\baselineskip}
% \end{table}

\begin{table}[ht]
    \centering
    \caption{Success rate and avg. episode length of agents in simulation.}
    \label{tab:results_success_rate}
	\vspace{-0.25\baselineskip}
    \setlength{\tabcolsep}{4.5pt}
    \begin{tabular}{lcccc} \hline
        \textbf{Agent} & \textbf{T1} & \textbf{T1(n)} & \textbf{T2} & \textbf{T2(n)} \\ \hline
        Teacher         & \textbf{92.6\%}\textit{(36.9s)} & 77\%\textit{(67.0s)} & \textbf{87\%}\textit{(34.0s)} & 54\%\textit{(69.5s)} \\
        Teacher\textsubscript{noise} & 69\%\textit{(42.5s)} & 66\%\textit{(33.7s)}  & 46\%\textit{(33.2s)} & 43\%\textit{(31.5s)}  \\
        Student         & 91.4\%\textit{(40.9s)}  & \textbf{90\%\textit{(42.8s)}}  & 85\%\textit{(45.9s)} & \textbf{63\%}\textit{(42.7s)}  \\ \hline
    \end{tabular}
	\vspace{-0.8\baselineskip}
\end{table}

Comparing the teacher and student agent it can be observed that the teacher performs slightly better in all categories if no noise is present, but with noise, the teacher's performance significantly decreases as it processes all noise as if it were valid observations. As soon as noise is added, the student agent performs better. The addition of smaller stones further amplifies this effect. In this case, the student clearly outperforms the teacher in both success rate as well as average duration. 

\subsection{Zero-Shot Sim-to-Real Transfer}

The physical rover is used to test teacher and student agent in a real-world environment. The rover is placed in the same position and given the same target for both agents to reach. % When it has reached the target, it will be reset, and the new target is the previous starting position.
%To test if the sim 2 real gab has been decreased, experiments are made both in simulation and in real-life, and the behaviors from the two are compared.
% Maybe this should not be included? - Thoughts are welcome:-)
Initially, the agents are tested in a flat terrain with obstacles, but they have difficulty navigating around obstacles that are directly in front of them as they try to reach their goal. As a result, tests with obstacles are deemed unsuccessful, although the agents seem to do better when the obstacles are slightly to the side.

\newcommand{\greenline}{\raisebox{2pt}{\tikz{\draw[-,black!40!green,solid,line width = 1.5pt](0,0) -- (4mm,0);}}}
\newcommand{\redline}{\raisebox{2pt}{\tikz{\draw[-,black!40!red,solid,line width = 1.5pt](0,0) -- (4mm,0);}}}

\begin{figure}[ht]
    \vspace{0mm}
    \centering
    \includegraphics[width=1.0\linewidth]{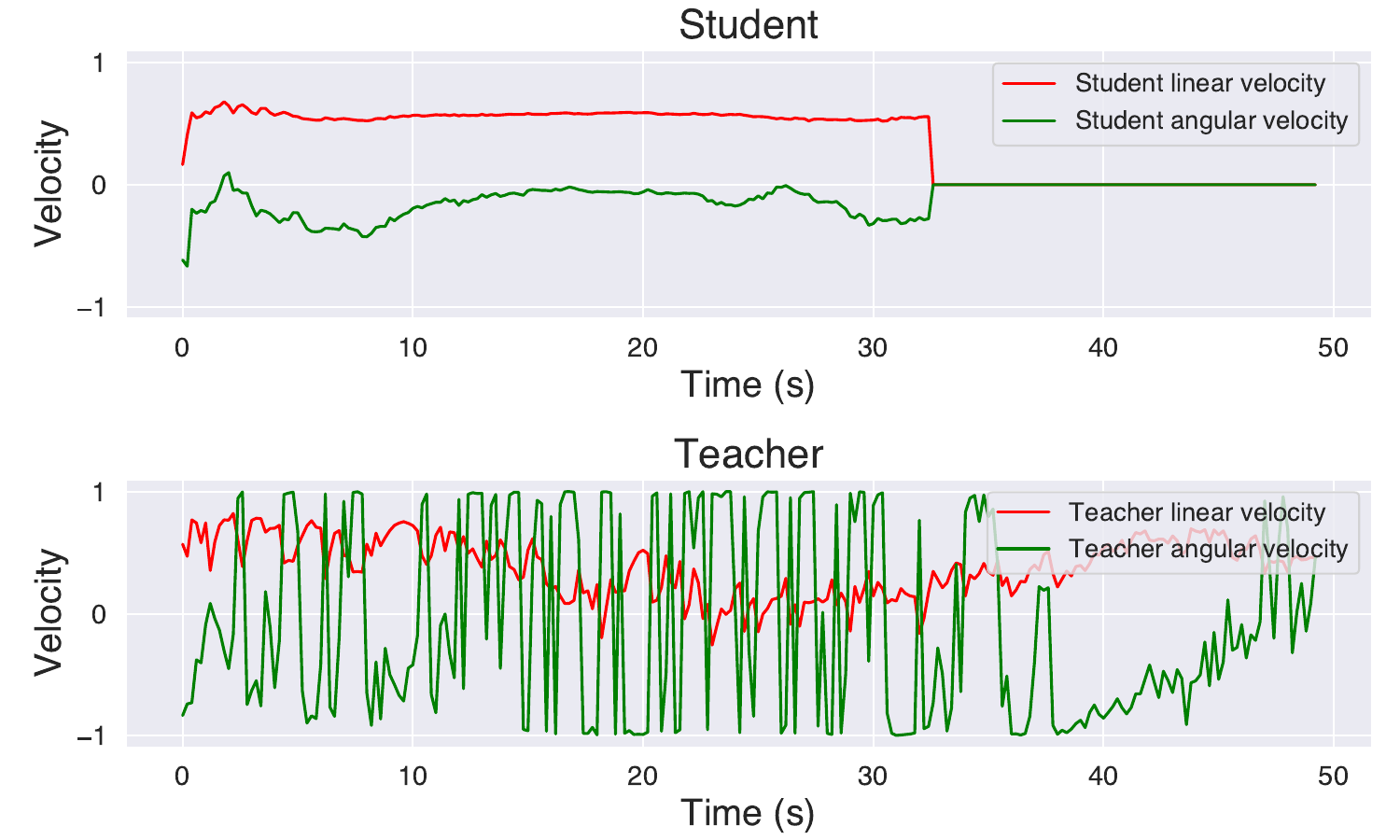}
    \vspace*{-5mm}
    \caption{The student and teachers performance in angular velocity \protect\greenline \phantom{-}and linear velocity \protect\redline \phantom{-} over time. The student oscillates less on the output actions and finishes faster.}
    \label{fig:Test:Angular}
\end{figure}

Hence, both agents are tested in a flat environment without obstacles, to see how they perform in the presence of noise from a physical sensor. Five tests are conducted with each agent, and the result of two of the runs are visualized in \autoref{fig:Test:Angular}.

% \section{Discussion}
The results of the experiments show that the student does not only reaches the goal more rapidly than the teacher but also exhibits fewer oscillations in its outputs when approaching a goal point. This difference arises because the teacher struggles to manage the noise inherent in real-life systems, whereas the student, having been trained to mitigate some of this noise, performs more adeptly. Moreover, when tested with noise in simulation, the student achieves a higher success rate and arrives at the goal more swiftly. These findings underscore the potential of our proposed method in bridging the reality gap.

The student agent's behavior in real-world scenarios closely mirrors its performance in simulations, more so than the teacher agent. The pronounced oscillations observed in the teacher's action outputs in real-world settings, attributed to noise, are substantially diminished with our method. Thus, it is evident that our two-stage learning strategy effectively supports the sim-to-real transferability of the agent.

\section{Conclusion and future works}
\label{sec:conclusion}
In conclusion, we demonstrate a novel two-stage training paradigm aimed at mitigating the reality gap in extraterrestrial mapless navigation for space rovers. We transitioned the refined student policy to a physical space rover, mirroring the kinematic configuration of NASA's Perseverance, and conducted tests to assess the impact of our method on the reality gap. The outcomes reveal that the student policy is notably more resilient to noise, both in simulations and in real-world settings, and adeptly navigates to designated targets. Furthermore, the student policy's behavior in real-world tests closely aligns with its simulated performance, suggesting our approach significantly enhances sim-to-real transfer. This method, rooted in a two-stage learning process, holds promise in narrowing the reality gap. 
Broadly speaking, we are optimistic about the potential of reinforcement learning techniques in augmenting the autonomy of space exploration rovers. 
In the future, we will investigate the impact of diverse noise types on teacher observations and the potential benefits of fine-tuning noise parameters. We also aim to optimize the student network's architecture and parameters for enhanced performance. A crucial task will be aligning the rover's dynamics in simulations more closely with its real-world counterpart to ensure more accurate policy transfers.

%%%%%%%%%%%%%%%%%%%%%%%%%%%%%%%%%%%%%%%%%%%%%%%%%%%%%%%%%

\addtolength{\textheight}{-0cm}  

%%%%%%%%%%%%%%%%%%%%%%%%%%%%%%%%%%%%%%%%%%%%%%%%%%%%%%%%%

\bibliographystyle{bibliography/_style/IEEEtran.bst}
\bibliography{bibliography/bibliography.bib}

\end{document}